\documentclass{bmvc2k}

\usepackage{booktabs}
\usepackage{amsfonts, amsmath}
\usepackage{floatrow}
\newfloatcommand{capbtabbox}{table}[][\FBwidth]
\usepackage{hyperref} 

\usepackage{blindtext}

\newcommand{\namecerfspell}[1]{CER\textsubscript{fspell}}
\newcommand{\namecerfull}[1]{CER\textsubscript{full}}
\newcommand{\modelname}[1]{Transpeller}
\usepackage{graphicx}

\newcommand\blfootnote[1]{%
  \begingroup
  \renewcommand\thefootnote{}\footnote{#1}%
  \addtocounter{footnote}{-1}%
  \endgroup
}

\usepackage[subtle]{savetrees}
\def\sepappendix{0}

\title{Weakly-supervised Fingerspelling Recognition in British Sign Language Videos}

\addauthor{K R Prajwal*}{http://robots.ox.ac.uk/~prajwal}{1}
\addauthor{Hannah Bull*}{https://hannahbull.github.io}{2}
\addauthor{Liliane Momeni*}{https://www.robots.ox.ac.uk/~liliane}{1}
\addauthor{Samuel Albanie}{https://samuelalbanie.com}{3}
\addauthor{Gül Varol}{http://imagine.enpc.fr/~varolg}{4}
\addauthor{Andrew Zisserman}{http://robots.ox.ac.uk/~az}{1}

\addinstitution{
 Visual Geometry Group\\
 Department of Engineering Science\\
 University of Oxford\\
 Oxford, UK
}
\addinstitution{
 LISN, Univ Paris-Saclay, \\
 CNRS, France
}
\addinstitution{
 Department of Engineering, 
 University of Cambridge, UK
}
\addinstitution{
 LIGM, École des Ponts, 
 Univ Gustave Eiffel, 
 CNRS, France
}

\runninghead{Prajwal et al.}{Weakly-supervised Fingerspelling Recognition in BSL Videos}

\begin{document}

\maketitle

\blfootnote{$^\ast$ Equal Contribution.}

%\vspace{-20pt}

\vspace{-20pt}
\begin{abstract}
The goal of this work is to detect and recognize sequences of letters signed using fingerspelling in British Sign Language (BSL). Previous fingerspelling recognition methods have not focused on BSL, which has a very different signing alphabet (e.g., two-handed instead of one-handed) to American Sign Language (ASL). They also use manual annotations for training. In contrast to previous methods, our method only uses weak annotations from subtitles for training. We localize potential instances of fingerspelling using a simple feature similarity method, then automatically annotate these instances by querying subtitle words and searching for corresponding mouthing cues from the signer. We propose a Transformer architecture adapted to this task, with a multiple-hypothesis CTC loss function to learn from alternative annotation possibilities. We employ a multi-stage training approach, where we make use of an initial version of our trained model to extend and enhance our training data before re-training again to achieve better performance. Through extensive evaluations, we verify our method for automatic annotation and our model architecture. Moreover, we provide a human expert annotated test set of $5K$ video clips for evaluating BSL fingerspelling recognition methods to support sign language research. 
\end{abstract}

\begin{figure}[htb]
    \centering
    \includegraphics[width=\textwidth]{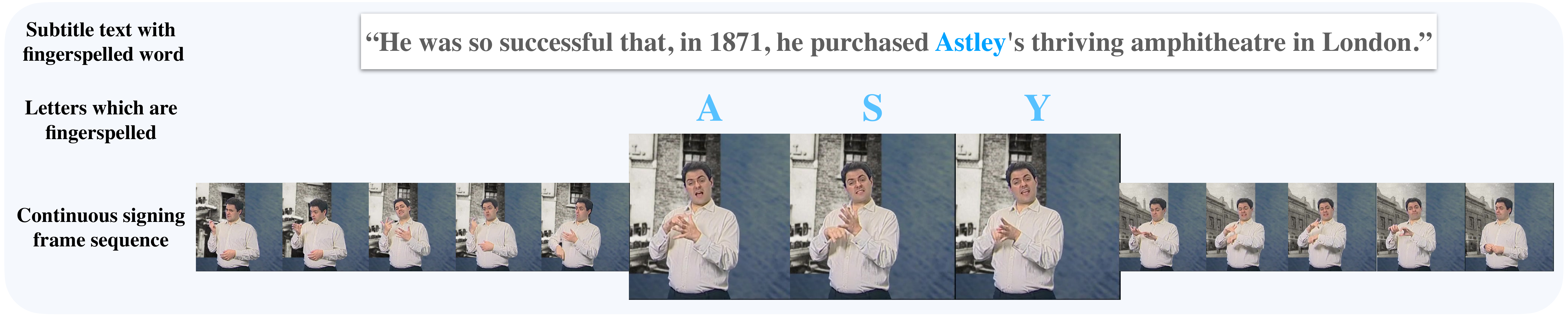}
    \vspace{-0.6cm}
    \caption{\textbf{Fingerspelling recognition.} We study the task of recognising
    % from scratch \gul{from scratch?}
    a sequence of BSL fingerspelled letters in a continuous signing window in sign language interpreted TV broadcast data. We exploit the accompanying English subtitles to automatically collect training data. However, the task remains very challenging as (i) the signer may fingerspell only a subset of the letters -- as shown above, although the subtitle contains the word `Astley', only the letters `A', `S' and `Y' are fingerspelled -- and (ii) due to occlusions from the hands, different letters can visually look very similar.
    %: `A', `S', `Y' for example in this case.
    }
    \label{fig:teaser}
\end{figure}

\vspace{-0.6cm}
\section{Introduction}
\label{sec:intro}
\vspace{-0.1cm}

%%%% IMPORTANCE OF FINGERSPELLING:
Fingerspelling in signed languages is a means to encode words from the surrounding written language into sign language via a manual alphabet, i.e.\ one sign per letter. Words from a written language with no known sign may be fingerspelled, such as names of people and places. Additionally, some signs are derived from fingerspelled words, for example, initialized signs in BSL such as `F' for `father' or `KK' for `kitchen' \cite{sutton1990variation}. Padden~\&~Gunsauls~\cite{padden2003} estimate that signers fingerspell 12-35\% of the time in ASL. Within the BSL videos used in this work~\cite{albanie21bobsl}, we estimate roughly 5-10\% based on the
duration of automatically detected fingerspelling content.
% frequency of proper noun occurrences.
Consequently, it is important to incorporate automatic fingerspelling recognition methods to be able to \textit{exhaustively} transcribe signs in continuous sign language videos.
% \hannah{will integrate \cite{hanson1981word} % Results indicated that these deaf subjects did not read fingerspelled words as individual letters. Rather, subjects made use of the underlying structure of words.
%Research on one-handed fingerspelling indicates that it is recognition of the overall pattern of movement that enables comprehension, rather than the sequence of the individual manual letters themselves (Hanson 1981; Akamatsu 1985)
% \cite{nicodemus2017cross} %Partially fingerspell words statistics
% }

% THE CHALLENGE \az{add here also the challenge of occlusions and segmentation when two hands are used}

Some sign languages, such as ASL and LSF, use a one-handed manual alphabet, and others, including BSL and Auslan, use a two-handed manual alphabet \cite{schembri2007sociolinguistic}.
One-handed manual alphabets typically do not involve significant wrist movements, while two-handed fingerspelling resembles other lexical signs, making it relatively more difficult to detect fingerspelling beginning and end times from a longer signing sequence.
The presence of two-hand movements further results in occlusions, making it challenging to differentiate between certain characters, e.g.\ especially vowels. 

%%%% TASK DEFINITION, BRIEFLY WHAT WE DO:
In this work, we focus on BSL, i.e.\ the more challenging two-handed fingerspelling.  Given a sign language video, our goal is to temporally locate the fingerspelling segments within the video (i.e.\ detection) and to transcribe the fingerspelled letters to text (i.e.\ recognition). To this end, we design a Transformer-based model that jointly performs both detection and recognition. Our key contribution lies in the data collection procedure used to automatically obtain training data for this task, which is applicable to any sign language videos that have approximately-aligned subtitle translations. We also provide the first large-scale benchmark for BSL fingerspelling recognition based on the recently released BOBSL dataset~\cite{albanie21bobsl}.
Our experiments on this benchmark demonstrate promising results with a 53.3 character error rate on this challenging task while only using weak supervision.

%%%% AUTOMATIC ANNOTATIONS, CONTRASTING TO PREVIOUS WORK WITH MANUAL
Previous works building fingerspelling datasets rely on manual annotation, either by
expert
% in house
annotators resulting in limited data~\cite{shi2018american}, or by crowdsourcing noisy large-scale annotations~\cite{shi2019fingerspelling}. In contrast to these works on ASL fingerspelling, we introduce a practical
methodology to automatically annotate fingerspelling in the presence of subtitled sign language
video data, allowing to scale up the data size, and potentially to be applicable to other sign languages. Starting from a small number of manually annotated fingerspelling exemplars, we use an embedding space to find numerous similar instances of fingerspelling in the corpus. To annotate these instances, we exploit the observation that signers often simultaneously mouth the words
(i.e.\ silent speech with lip movements)
which they fingerspell. We obtain an initial set of annotations by querying potential words from the subtitles, especially proper nouns, and identifying mouthing cues~\cite{albanie2020bsl,prajwal2021visual} which also coincide with fingerspelling instances.
This initial set of
annotations
are further extended and enhanced by a pseudolabeling step, also making use of subtitles
(see Sec.~\ref{subsec:annotations}).

%%%% MH-CTC CONTRIBUTION
A key challenge with training through automatic annotations is label noise. For example, the mouthing model~\cite{prajwal2021visual} may spot a single word, while the fingerspelling contains multiple words, such as name-surname pairs.
In some other cases, the mouthing model may fail, associating the wrong word in the subtitle with the fingerspelling segment. Also,
 some letters of the word may be skipped within fingerspelling as illustrated in Fig.~\ref{fig:teaser}.
To account for this uncertainty, we implement a multiple-hypotheses version of the CTC loss~\cite{graves2006connectionist} (MH-CTC)
where we consider all nouns from the subtitle, as well as bigrams and trigrams, as potential targets. 
% Finally, notice that our method of using words from subtitles introduces noise as some letters may be skipped within fingerspelling (see Fig.~\ref{fig:teaser}). We adopt a random character dropping strategy during training to take this into account.
% \az{Should explain that our method of using words from subtitles introduces noise as not all letters signed. Then flesh this out} 
% We notice that the temporal boundaries tend to be of relatively high quality, compared to the associated characters? Not sure.
%, how fingerspelling can cover partial words => maybe not if we do not want to emphasize too much dropping characters (if the improvement is small)

%%%% CONTRIBUTIONS SUMMARY:
Our main contributions are: 
(i)~Training a BSL fingerspelling detection and recognition model using only weak labels from subtitles, mouthing cues, and a small number (115) of manual fingerspelling exemplars;
(ii)~Employing multiple hypotheses from the subtitle words within the CTC loss to train with noisy labels;
(iii)~Demonstrating advantages of a pseudolabeling step incorporating the subtitle information; 
and
(iv)~Providing a large-scale manually annotated benchmark for evaluating BSL fingerspelling recognition, released for research purposes. Please check our website for more details: \url{https://www.robots.ox.ac.uk/~vgg/research/transpeller}.
\vspace{-0.3cm}
\section{Related work}
\label{sec:related}
\vspace{-0.2cm}

Our work relates to four themes in the research literature: the broader topics of \textit{sign language recognition} and \textit{learning sign language from weak/noisy annotation}, and the more directly related literature on \textit{spotting mouthings} and \textit{fingerspelling recognition}.
% We discuss connections to each theme next.\\

% \az{Should also mention methods for text spotting that both detect words/sentences (e.g.\ using Faster-RCNN) and also read them using, e.g.\ CTC, in the classification  head.}
% \samuel{Did you have a particular reference in mind? I found this one~\cite{borisyuk2018rosetta}, which user Faster-RCNN together with CTC for text detection and recognition. We could cite this, but the architecture is fairly distant from ours (it uses two systems that are trained independently).}
\noindent\textbf{Sign language recognition.} 
Building on the pioneering 1988 work of Tamura and Kawasaki \cite{tamura1988recognition}, early approaches to automatic sign recognition made use of hand-crafted features for motion~\cite{yang2002extraction} and hand shape~\cite{fillbrandt2003extraction,vogler2003handshapes}.
To model the temporal nature of signing, there has also been a rich body of work exploring the use of Hidden Markov Models~\cite{starner1995visual,vogler2001framework,fang2004novel,cooper2011reading,koller2016deep,koller2017re} and Transformers~\cite{camgoz2020sign,de2020sign}.
One notable trend in prior work is the transition towards employing deep spatiotemporal neural networks to provide robust features for recognition.
In this regard, the I3D model of~\citet{carreira2017quo} has seen widespread adoption, achieving strong recognition results on a range of benchmarks~\cite{Joze19msasl,Li19wlasl,albanie2020bsl,li2020transferring}.
In this work, we likewise build our approach on strong spatiotemporal video representations, adopting the Video-Swin Transformer~\cite{liu2022video} as a backbone for our model.

\noindent\textbf{Learning sign language from weak/noisy annotation.} 
Given the paucity of large-scale annotated sign language datasets, a range of prior work has sought to leverage weakly aligned subtitled interpreter footage as a supervisory signal~\cite{cooper2009learning,buehler2009learning,pfister2014domain,Momeni20b} for sign spotting and recognition via apriori mining~\cite{agrawal1993mining} and multiple instance learning~\cite{dietterich1997solving}.
Similarly to these works, we likewise aim to make use of subtitled signing footage.
However, we do so
%(in combination with other weak cues obtained via mouthing cues~\cite{pfister2013large,albanie2020bsl}),
in order to detect and recognize fingerspelling in a manner that allows for scalable training.
To the best of our knowledge, this approach has not been considered in prior work.

\noindent \textbf{Spotting mouthings in sign language videos.} Signers often mouth the words that they sign (or fingerspell)~\cite{sutton2007mouthings}. The recent advancements in visual keyword spotting~\cite{prajwal2021visual,stafylakis2018zero,momeni2020seeing} have enabled the automatic curation of large-scale sign language datasets by spotting a set of query words using the mouthing cues and matching them with the corresponding sign segment. The state-of-the-art architecture for the visual KWS task is the Transpotter~\cite{prajwal2021visual}, which we use in Sec~\ref{subsubsec:exemplarannots} to obtain our initial set of automatic fingerspelling annotations. 
Our fingerspelling architecture also partly takes inspiration from the Transpotter
(and more broadly from prior works for text spotting that detect words and learn to read them via CTC~\cite{li2017towards,borisyuk2018rosetta}), 
wherein we process video features with a single Transformer encoder and then employ multiple heads to solve related tasks such as detection and classification with a $[CLS]$ token. 
In this work, we add also add a recognition head supervised by a novel loss function. We also show the benefits of our multi-stage training pipeline in training this model when we only have weak supervision.    
% Fingerspelling recognition is fundamentally a sequence recognition task.
% However, our sequences are noisily labelled.
% We build on CTC~\cite{graves2006connectionist}.

\noindent\textbf{Fingerspelling recognition and detection.} 
% mention that previous work also jointly does detection+recognition
Early work on automatic fingerspelling recognition explored the task of classification under fairly constrained settings, focusing on isolated signs and limited vocabularies (e.g.\ 20 words~\cite{goh2006dynamic}, 82 words~\cite{ricco2009fingerspelling} and 100 words \cite{liwicki2009automatic}).
Kim et al. propose to consider instead a ``lexicon-free'' setting
% to lift this limitation and introduce a dataset
which they tackle with frame-level classifiers in combination with segmental CRFs on a newly introduced dataset, of 3,684 American Sign Language (ASL) fingerspelling instances~\cite{kim2017lexicon}.
Moving towards more challenging data, the ChicagoFSWild (7304 fingerspelling sequences across 160 signers annotated by ASL students)~\cite{shi2018american} and ChicagoFSWild+ (55,232 fingerspelling sequences signed by 260 signers annotated by crowdsourcing)~\cite{shi2019fingerspelling} datasets sourced from YouTube and Deaf social media target greater diversity and visual variation.
% across four signers annotated with the time of peak articulation for each fingerspelled letter
% to study this setting~\cite{kim2017lexicon}.
% This work demonstrates the utility of segmental CRFs for this task in combination with frame-level deep neural network classifiers, highlighting the challenges of signer-independent recognition.

% -Lexicon-free fingerspelling recognition from video: Data, models, and signer adaptation (Kim et al., 2017)~\cite{kim2017lexicon}
From a modeling perspective, Pugealt and Bowden employ random forests on depth and intensity images for real-time recognition of 24 fingerspelled letters~\cite{pugeault2011spelling}.
Shi et al. demonstrate the benefits of using a signing hand detector for fingerspelling recognition without frame-level labels~\cite{shi2017multitask}, motivating later work to attain this benefit automatically through visual attention without an explicit region detector~\cite{shi2019fingerspelling}.
Other work has explored the feasibility of using synthetic hand training data to fine-tune a CNN for isolated Irish Sign Language (ISL) fingerspelling recognition~\cite{fowley2021sign}.
% -Multitask training with unlabeled data for end-to-end sign language fingerspelling recognition (Shi et al., ASRU 2017)~\cite{shi2017multitask}

More closer to our work, several works have considered detecting the temporal location of fingerspelling in addition to recognition.
This includes efforts to segment signing into sign types (classifiers, lexical signs, and fingerspelling) prior to recognition~\cite{yanovich2016detection}, as well as systems for fingerspelling detection supervised with segment boundaries~\cite{shi2021fingerspelling}.
% -Fingerspelling Detection in American Sign Language (Shi et al., CVPR 2021)~\cite{shi2021fingerspelling}
The recently proposed FSS-Net learns joint embeddings to enable fingerspelling search within and across videos~\cite{shi2022searching}.
In contrast, our work is weakly supervised with noisy, automatic annotations.
It is weakly supervised in the sense that the model lacks access to ground truth fingerspelling boundaries at training, while the annotation is noisy in the sense that it is derived from subtitles from which the signing is produced as a translation, rather than a transcription.\\

\vspace{-1cm}
\section{Fingerspelling detection and recognition}
\label{sec:method}
\vspace{-0.2cm}
In this section, we introduce Transpeller, our Transformer-based model to recognize and detect fingerspelling (Sec.~\ref{subsec:model}). Our model is trained only on automatically curated data. In order to circumvent this label noise, we also propose a new loss function in Sec.~\ref{subsec:mhctc}. 

\vspace{-0.3cm}
\subsection{The \modelname{} architecture}
\label{subsec:model}
\begin{figure}[t]
    \centering
    \includegraphics[width=\textwidth]{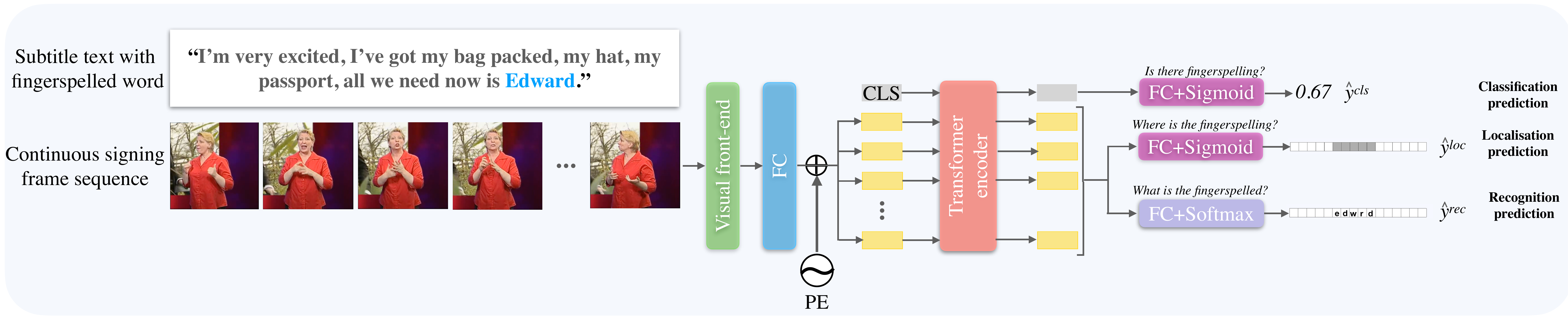}
    \vspace{-0.6cm}
    \caption{\textbf{Transpeller architecture.} Given a short clip of a signer, we extract features from a visual front-end pre-trained on the sign language recognition task. We project these features to a desired hidden dimension and add positional encodings before encoding these feature vectors using a transformer encoder. We use three heads on top of the transformer outputs to predict if the clip contains fingerspelling (classification prediction), and if so, where it is in the clip (localization prediction) and what it is (recognition prediction). The subtitle shown on the left is used for curating the training data and is shown here for illustrative purposes. 
    }
    \label{fig:model}
\end{figure}

\vspace{-0.2cm}
Our model ingests a video clip of a signer, encodes it with a Transformer encoder~\cite{vaswani2017attention}, and produces three outputs with each of its prediction heads: (i) a classification head that predicts \textbf{if} the given video clip contains fingerspelling, (ii) a localization head that produces per-frame probabilities indicating \textbf{where} the fingerspelling is in the clip, (iii) a recognition head that produces a sequence of letter probabilities indicating \textbf{what} is being fingerspelled. We illustrate this architecture in Figure~\ref{fig:model}.

\noindent\textbf{Visual backbone.} Our input is a sequence of RGB frames constituting a short clip of a signer. In order to extract the visual features, we follow prior sign-language works~\cite{bull2021aligning,Momeni22,varol2021read} and use a pre-trained sign classification model. The sign classification model used in prior works is an I3D model~\cite{bull2021aligning,Momeni22,varol2021read} pre-trained on Kinetics~\cite{kay2017kinetics} and finetuned using a sign classification objective. We replace the I3D with a relatively modern Video-Swin-S~\cite{liu2022video} architecture and finetune it for the sign classification task in a similar fashion. We pre-extract the features for all our videos and save them at a temporal stride of 4. The Transpeller operates on these feature vectors $V \in \mathbb{R}^{T \times d_f}$.

\noindent\textbf{Transformer Encoder.} Given a sequence of visual feature vectors $V \in \mathbb{R}^{T \times d_f}$, we first use two fully-connected layers ($FC$) to project the feature dimension to $d$, which is the hidden dimension of the transformer encoder. We add the temporal positional encodings and prepend this sequence with a learnable $[CLS]$ token embedding (such as in BERT~\cite{devlin2018bert} and ViT~\cite{dosovitskiy2020image}). We encode the temporal information of this sequence by passing it through a Transformer encoder consisting of $N$ layers:  
\vspace{-0.2cm}
\[
V_{enc} =
encoder([CLS]; [FC(V) + PE_{1:T}]) \in \mathbb{R}^{(1+ T) \times d}.
% $$
    \vspace{-0.2cm}
\]
\noindent\textbf{Prediction heads.} The $[CLS]$ output feature vector $V_{enc(1)}$ serves as a aggregate
representation for the entire input video clip. 
An MLP head for binary classification, $f_c$ is attached to $V_{enc(1)}$ to predict the probability of a fingerspelling segment being present in the input video:
\vspace{-0.2cm}
\[
% $$
\hat{y}^{cls} = \sigma( f_{c} ( V_{enc(1)} ) ) \in \mathbb{R}^{1},
% $$   
\vspace{-0.2cm}
\]
where $\sigma$ denotes a sigmoid activation.
To localize the fingerspelling segment, we attach a second MLP head $f_l$ that is shared across the encoded video feature time-steps $V_{enc(2:T + 1)}$:
% \vspace{-5pt}
\vspace{-0.2cm}
\[
% $$
\hat{y}^{loc} = \sigma( f_{l} ( V_{enc(2:T + 1)} ) ) \in \mathbb{R}^{T}.
% $$
    \vspace{-0.2cm}
\]
The output $y^{loc}_t$ at each feature time-step $t \in T$ indicates the probability of the time-step $t$ being a part of a fingerspelling segment. In order to recognize what is being fingerspelt, a third MLP head $f_r$ is attached in a similar fashion to $f_l$ to predict $\mathbb{C}$ letter probabilities at each time-step:
\vspace{-1pt}
\[
% $$
\hat{y}^{rec} = softmax( f_{r} ( V_{enc(2:T + 1)} ) ) \in \mathbb{R}^{T \times \mathbb{C}}.    \vspace{-0.3cm}
% $$
\]

\noindent\textbf{Loss functions.}
Given a training dataset $\mathcal{D}$ consisting of video clips $v$, class labels $y^{cls}$, location labels $y^{loc}$ and recognition labels (whenever present) $y^{rec}$, we define the following training objectives:
\vspace{-1pt}
\begin{align}
\mathcal{L}^{cls}
&=
-
\mathbb{E}_{(v, y^{cls}) \in \mathcal{D}} \
% \left[
BCE(y^{cls},\hat{y}^{cls})
% \right]
\label{eqn:loss_cls}
% \end{align}
\\
% \begin{align}
\mathcal{L}^{loc}
&=
-
\mathbb{E}_{(v, y^{cls}, y^{loc}) \in \mathcal{D}} \
y^{cls}
\left[
\frac{1}{T}
\sum_{t=1}^{T}
BCE(y^{loc}_t ,\hat{y}^{loc}_t )
\right]    \vspace{-0.3cm}
\label{eqn:loss_loc}
% \end{align}
\\
% \text{where} \\
% \begin{align}
% BCE(y,\hat{y})
% &=
% y \log \hat{y} + (1-y) \log (1-\hat{y}),
% \label{eqn:bce}
% \\
\mathcal{L}^{rec}
&=
-
\mathbb{E}_{(v, y^{rec}) \in \mathcal{D}} \
CTC(y^{rec}, \hat{y}^{rec}) 
\label{eqn:loss_rec}     \vspace{-0.4cm}
\end{align}
% \begin{align}
% \mathcal{L}^{cls}
% &=
% -
% \mathbb{E}_{(v, y^{cls}) \in \mathcal{D}} \
% % \left[
% BCE(y^{cls},\hat{y}^{cls})
% % \right]
% \label{eqn:loss_cls}
% % \end{align}
% \\
% % \begin{align}
% \mathcal{L}^{loc}
% &=
% -
% \mathbb{E}_{(v, y^{cls}, y^{loc}) \in \mathcal{D}} \
% y^{cls}
% \left[
% \frac{1}{T}
% \sum_{t=1}^{T}
% BCE(y^{loc}_t ,\hat{y}^{loc}_t )
% \right]
% \label{eqn:loss_loc}
% \\
% \mathcal{L}^{rec}
% &=
% -
% \mathbb{E}_{(v, y^{rec}) \in \mathcal{D}} \
% CTC(y^{rec}, \hat{y}^{rec}) 
% \label{eqn:loss_rec} 

% % \text{where} \\
% % \begin{align}
% \end{align}
\noindent{}where BCE stands for the binary cross-entropy loss and CTC stands for Connectionist Temporal Classification~\cite{graves2006connectionist} loss. The labels $y^{cls}$ are set to $1$ when the given keyword occurs in the video and 0 otherwise;
the frame labels $y^{loc}$ are set to 1 for the frames where the keyword is uttered and 0 otherwise.
We optimize the total loss
% \begin{align}
$
\mathcal{L}
= \mathcal{L}^{cls} + \mathcal{L}^{loc} + \lambda \mathcal{L}^{rec}
\label{eqn:loss}
% \end{align}
$
where, $\lambda = 1$ if a given input video has a word label annotation, else, $\lambda = 0$ to only train the classification and localization heads. 

\vspace{-0.3cm}
\subsection{Multiple Hypotheses (MH) CTC loss}
\vspace{-0.2cm}
\label{subsec:mhctc}
As described earlier, our word labels for fingerspelling recognition are weak labels from subtitles and mouthing cues. The process for obtaining these labels is described in Sec.~\ref{subsubsec:exemplarannots} \&~\ref{subsubsec:modelannots}. Since this process is automatic, it introduces a degree of label noise. For example, the mouthing model can detect false positives; the detection boundaries can be erroneous, or multiple mouthings can be spotted for a single detection interval. We observed that our automatic detection pipeline is more accurate than our process of obtaining word labels. If we assume the fingerspelling detection is correct, then it is quite likely that the fingerspelled word is among one of the words (usually a noun) in the subtitle.

With this idea in mind, we design an improved CTC loss function, termed Multiple Hypotheses CTC (MH-CTC), that allows the model to ``pick'' the most correct word label from a set of possible word hypotheses. 
This set comprises a number of words, of which one is the correct word label for the fingerspelling sequence in the input video clip. 
For example, the set of word hypotheses could be the proper nouns in the subtitle corresponding to the input video clip. Given a list of word hypotheses $\mathbb{H}$ for the input video clip $v$, our modified recognition loss $\mathcal{L}^{rec}$ is: 
\vspace{-0.2cm}
\[
\mathcal{L}^{rec} = 
-\mathbb{E}_{(v, \mathbb{H}) \in \mathcal{D}} \
\min_{\forall h \in \mathbb{H}} CTC(h, \hat{y}^{rec})     \vspace{-0.3cm}
\]
This corresponds to backpropagating the recognition loss for the word that achieves the minimum CTC loss among all the hypotheses. Since this allows the model to ``choose'' its own target, we found two strategies that help the model converge. Firstly, pretraining the model with CTC loss using Eqn~\ref{eqn:loss_rec} before using MH-CTC is essential. Secondly, when using MH-CTC, we found that randomly (with 50\% chance) setting $\mathbb{H}$ to be a single hypothesis containing the word found by our automatic annotation also prevents the model from diverging.  

\vspace{-0.3cm}
\section{Automatic annotations}
\label{subsec:annotations}
\vspace{-0.2cm}
Our model is completely trained with weak labels that are automatically curated. We perform two stages of automatic annotation. 
In the first stage (Sec.~\ref{subsubsec:exemplarannots}), we automatically annotate fingerspelling detections using exemplars and letter labels using mouthing cues. In the second stage (Sec.~\ref{subsubsec:modelannots}), we obtain detection and letter pseudolabels from the \modelname{} model that has been trained on annotations from the first stage. The number and duration of these fingerspelling detections, as well as the number of detections associated with either a noun or proper noun letter label, is shown in  Tab.~\ref{tab:cosinesimilaritystats}.

\vspace{-0.3cm}
\subsection{Exemplar and mouthing annotations}
\label{subsubsec:exemplarannots}
\vspace{-0.2cm}
\noindent\textbf{Detections using exemplars.} We manually annotate a small number $E$ of single frame exemplars of fingerspelling amongst videos from different signers in the training set ($E=115$). Using these exemplars, we search for frames in continuous signing with high feature similarity to these frames containing fingerspelling.
This exemplar-based annotation technique is inspired by \cite{Momeni20b,Momeni22}.
We use features from \cite{Momeni22} and compute cosine similarity with fingerspelling features. This simple method provides approximate annotation of fingerspelling detections. We use this method for two reasons: firstly, to help our model learn fingerspelling temporal detection, and secondly, to select video segments containing fingerspelling for manual annotation. Technical details on the computation of the feature similarity can be found in the supplementary material. 
% \gul{let's choose between detection/segmentation/localisation} 

\begin{table} % [h]
    \centering
\resizebox{0.99\linewidth}{!}{
    \begin{tabular}{lccc|ccc}
        \toprule
         & \multicolumn{3}{c|}{Train} & \multicolumn{3}{c}{Val} \\
        & \#labels & vocab & avg. dur. & \#labels & vocab & avg. dur. \\

        \midrule
        Stage 1: Exemplar detections & 149k & - & 1.4s & 3.0k & - & 1.3s \\
        ~ w/ mouthing labels: nouns & 59k & 18k & 1.8s & 1.2k & 0.8k & 1.6s \\
        ~ w/ mouthing labels: pr.~nouns  & 39k & 14k & 1.9s & 0.9k & 0.5k & 1.7s \\ \hline
        Stage 2: \modelname{} detections & 129k & - & 1.9s & 2.5k & - & 1.8s \\
        ~ w/ mouthing labels: nouns &  61k & 19k & 2.1s & 1.3k & 0.9k & 1.9s \\
        ~ w/ mouthing labels: pr.~nouns & 41k & 15k & 2.2s & 0.9k & 0.5k & 2.0s \\
        ~ w/ \modelname{} labels  & 111k & 32k & 2.0s & 2.2k & 1.4k & 1.8s \\
        \bottomrule
    \end{tabular}
    }
    \caption{\textbf{Two stages of automatic annotations.} Stage 1: We use exemplars to detect fingerspelling and mouthing cues to obtain letter labels. Stage 2: We obtain detections and letter labels using \modelname{} pseudolabels. 
    }
    \vspace{-0.3cm}
    \label{tab:cosinesimilaritystats}
\end{table}

\noindent\textbf{Letter labels from mouthings.}
To obtain word annotations for the fingerspelled segments, we use mouthing cues, as fingerspelled words are often mouthed simultaneously in interpreted data~\cite{davis1990linguistic}. In~\cite{Momeni22}, the authors use an improved Transpotter architecture~\cite{prajwal2021visual} to query words from surrounding subtitles and localize corresponding mouthing cues. We consider all mouthing annotations from~\cite{Momeni22} falling within the interval of the fingerspelling detections. As the fingerspelling detection boundaries are approximate, and the automatic mouthing annotations are not always accurate, these annotations are noisy. Tab.~\ref{tab:test_set_annots} shows that almost all fingerspelling annotations refer to nouns, and most refer to proper nouns. Thus, restricting mouthing annotations to nouns or proper nouns reduces noise. %Tab.~\ref{tab:cosinesimilaritystats} shows the number of automatic fingerspelling annotations when restricting to nouns or proper nouns. 

\vspace{-0.3cm}
\subsection{\modelname{} annotations}
\label{subsubsec:modelannots}
\vspace{-0.2cm}
\noindent\textbf{Improving detections with \modelname{} pseudolabels.} The model described in Sec.~\ref{subsec:model} outputs a per-frame localization score, predicting the presence of fingerspelling. After training this model, we can improve upon the exemplar-based detections using pseudolabels. Given localisation scores $s_1,...,s_N$ for a window of $N$ frames, we consider that the window contains fingerspelling if $\mathrm{max}(s_1,...,s_N)> t_1$. We consider that a sub-interval $[i,j]$ ($1<=i<=j<=N$) of this window contains fingerspelling if $s_i,...,s_j>=t_2$, where $t_2<t_1$. To smooth the localization scores, we take the moving maximum score amongst $K$ consecutive scores. We let $t_1=0.7$, $t_2=0.3$ and $K=5$.

\noindent\textbf{Improving letter labels with \modelname{} pseudolabels.} Using pseudolabels from the model in Sec.~\ref{subsec:model}, we can also improve automatic letter label annotations. After decoding the CTC outputs using beam search, we can then compute a proximity score with words from neighboring subtitles, i.e.\ dist($w_1$, $w_2$), where $w_1$ is a subtitle word and $w_2$ is the output of beam search decoding. The proximity score is a variant of the Levenshtein edit distance, but where deletions are not heavily penalized. This is because words such as `Sarah Jane' can be reasonably fingerspelled as `SJ'. Details of this proximity score are provided in the supplementary material. We can use this proximity score to find the subtitle word most likely to be fingerspelled. %Tab.~\ref{tab:cosinesimilaritystats} contains statistics on the number of letter labels.

% \todo{put below in suppmat?}

% \begin{enumerate}
%     \item All repeated letters are removed before computation, e.g. (HARRY becomes HARY). 
%     \item Letters in $w_2$ that are not in $w_1$ are penalised and removed, e.g. (MARY, MERY) becomes (MARY, MRY) with a malus of +1. 
%     \item The proportion of letters of $w_1$ not in $w_2$ is added as a malus, e.g. (MARY, MRY) has a malus of +1/4.
%     \item Correct prediction of the first letter reduces the edit distance by 1, e.g. (MARY, MRY) has a bonus of -1, but (MARY, ARY) does not. 
%     \item Insertions and subtitions each carry a malus of +1, e.g. (MARY, MYR) has a malus of +1.
%     \item Deletions are not penalised.
% \end{enumerate}

\vspace{-0.3cm}
\subsection{Multi-stage training}
\label{subsec:training}
\vspace{-0.2cm}
We perform a multi-stage training strategy where we start by training on exemplar-based annotations (lines 2 and 3 of Tab.~\ref{tab:cosinesimilaritystats}) using the vanilla CTC loss to supervise the recognition head. Upon convergence, we finetune this model further using MH-CTC, this time additionally considering proper nouns in neighboring subtitles as the hypotheses. 

Using the above pre-trained model, we now extract the Transpeller annotations as detailed in Sec.~\ref{subsubsec:modelannots}. Using these new annotations, we repeat the process: we start with the vanilla CTC loss and then finetune this model further using MH-CTC. Given that we have pseudo-labels from the Stage 1 Transpeller model, we can restrict the MH-CTC search space. The hypotheses now only contain nouns and proper nouns from neighboring subtitles with at least one letter in common with the CTC decoded outputs from the \modelname{} model of Stage 1.

% Finally, we finetune this model by adding the pseudolabels to the training (see Sec.~\ref{subsec:annotations}).

% During training, we randomly drop letters to account for the noisy in the annotations that may cover partial letters from the ground truth word.

%Two training techniques for removing noise in automatic annotations: 1) restricting to nouns and proper nouns, 2) MH-CTC letter dropping. 

%\noindent\textbf{Restricting to nouns and proper nouns.} We restrict to nouns and proper nouns to evalutate the compromise between the number and quality of automatic annotations. 

%\noindent\textbf{Dropping letters as a data augmentation strategy} Not all letters are fingerspelled, \todo{cite}. We randomly drop letters during training and use these multiple hypotheses with CTC decoding. 

%\noindent\textbf{training schedule.}

\vspace{-0.3cm}
\section{Experiments}
\label{sec:experiments}

\vspace{-0.3cm}
\subsection{BOBSL Fingerspelling benchmark}
\label{subsec:benchmark}
\vspace{-0.2cm}
We collect and release the first benchmark for evaluating fingerspelling in British Sign Language. The test set annotations are collected by adapting the VIA Whole-Sign Verification Tool~\cite{dutta2019via} to the task of fingerspelling recognition. Given proposed temporal windows around
the automatic exemplar and mouthing annotations, annotators mark whether there is any fingerspelling in the signing window and type out the exact letters which are fingerspelled. We use a temporal window of 2.1s before to 4s after the midpoint of the automatically detected fingerspelling instance. Since the fingerspelled letters could be a subset of the actual full word (e.g.\ SH for SARAH), we also obtain the corresponding full word annotations. Descriptive statistics on the test set annotations are in Tab.~\ref{tab:test_set_annots}.

\noindent\textbf{Evaluation criteria.}
We measure the fingerspelling recognition performance using the Character Error Rate (CER), which provides a normalized count of the substitution/deletion/insertion errors in the predicted letter sequence when compared to the ground-truth sequence. We report two CERs for the two different ground-truth annotations we have for each clip: (i) \namecerfspell{} - ground-truth is the actual fingerspelled letters which, as mentioned before, may only be part of a word, and (ii) \namecerfull{} - ground-truth is the full word annotation to which the fingerspelling refers to. Given that our model(s) are trained only with automatic weakly-supervised full-word annotations (Sec.~\ref{subsec:annotations}), these two different CER scores can help us see if the model learns to pick up on the fingerspelled letters, rather than only relying on the mouthing cues or memorizing the full word annotations. 

\noindent\textbf{Implementation details.} We use a batch size of 32 and an initial learning rate of $5e^{-5}$, which is reduced to $1e^{-5}$ after the validation loss does not improve for 3 epochs. At test time, we decode with a beam width of 30.
%We choose the checkpoint with the lowest validation loss and evaluate it on the manually verified test benchmark that we will describe below. We use beam width of $50$ while decoding the character sequence at test time. 
More implementation details are provided in the supplementary material. 
%All our models are trained on the automatic annotations for the videos in the \textit{train} split of the BOBSL dataset. Similarly, for validation, we use the automatic annotations of the videos in the \textit{val} split. 
\begin{table} % [h]
    \centering
    \begin{tabular}{ccccccc}
        \toprule
        \#labels & (full word) vocab. &
        % Avg. len. & 
        \%nouns & 
        \%pr.~nouns & 
        \%full &
        avg. \% missing \\\midrule
        4923 & 3442 & 
        % 5.8 & 
        96\% & 74\% & 22\% & 34\% \\\bottomrule
    \end{tabular}
        \vspace{-0.3cm}
    \caption{\textbf{Statistics on test set annotations.} Most fingerspellings refer to proper nouns or nouns. Around 22\% of fingerspelling instances contain all letters of the encoded word, but on average 34\% of the letters of a word are omitted during fingerspelling. 
    }
    \label{tab:test_set_annots}
\end{table}

\vspace{-0.3cm}
\subsection{Results}
\label{subsec:results}
\vspace{-0.2cm}

\begin{table}
\centering
\begin{tabular}{llll}
\toprule
\textbf{Annotations} & \textbf{\# Recognition ex.} & \textbf{\namecerfspell} & \textbf{\namecerfull} \\
\midrule
Exemplars + Mouthings: proper nouns    & 39k  &  58.5    &   62.1    \\
Exemplars + Mouthings: nouns         & 59k   &  58.6    &   62.9    \\ 
%~ with letter dropping data augmentation & 59k & 58.5 & 65.3 \\
~ finetuned with MH-CTC   &   59k (avg. 4 hypoth.)  &   \textbf{57.6} & 64.3 \\
\bottomrule
\end{tabular}
    \vspace{-0.3cm}
\caption{\textbf{Stage 1: \modelname{} model with exemplar + mouthing supervision.} All rows use 149k fingerspelling detections for training. When assigning a word label for these detections from the subtitle, we choose to look at nouns, especially proper nouns, as they are more likely to be fingerspelt. 
%However, restricting only to proper nouns reduces the number of training samples significantly. 
We obtain the best results when using MH-CTC loss.}
\label{tab:exp_mouthing_annots}
\end{table}

\begin{table}
\centering
\begin{tabular}{llll}
\toprule
\textbf{Annotations} & \textbf{\# Recognition ex.} & \textbf{\namecerfspell} & \textbf{\namecerfull} \\ \midrule
 %Exemplars + Mouthings (line 2 Tab.~\ref{tab:cosinesimilaritystats})  &              59k            &     55.2    & \\
\modelname{} detect. + Mouthings  & 61k & 57.5 & 63.1 \\
\modelname{} detect. + Char. labels  & 111k & 55.4 & 63.0 \\ 
~ finetuned with MH-CTC  & 111k + (avg. 9 hypoth.) & \textbf{53.3} & \textbf{60.1} \\\bottomrule
\end{tabular}
    \vspace{-0.3cm}
\caption{\textbf{Stage 2: \modelname{} pseudolabels.} All rows use 129k fingerspelling detections for training. Using both the refined detections and word annotations from the Stage 1 model gives a clear reduction in the CER. Further, using MH-CTC gives a 2.1 CER boost. Overall, our Stage 2 achieves a final best CER of 53.3 which is 4.3 CER better than the best model of Stage~1, thus validating the impact of our multi-stage training pipeline.}
\label{tab:exp_pseudolabels}
\end{table}

We now evaluate different variations at each stage of the Transpeller recognition pipeline.

\noindent\textbf{Using exemplar detections and mouthing cue letter labels with CTC loss.} 
Our initial set of annotations from the exemplar detections gives us 149k fingerspelling instances, out of which a fraction of them can be associated with word labels using mouthing cues. In Tab.~\ref{tab:exp_mouthing_annots}, we show how our performance depends on our choice of recognition annotations. Restricting our automatic word label annotations to proper nouns gives us a cleaner training set, as they are most likely to be fingerspelt. However, this comes at the expense of having very few training samples. In row 2 of Table~\ref{tab:exp_mouthing_annots}, we find that we can tolerate a bit of label noise and expand to all nouns. %In row 3, we find that randomly dropping letters as an augmentation strategy helps us to combat overfitting owing to our dataset size. 
We finetune the best CTC-based model from the above using our MH-CTC loss, which further results in an improvement of 1.0 \namecerfspell{}. This is our best model with the initial set of annotations. 

\noindent\textbf{\modelname{} detections.}
We now expand our training data, by extracting pseudolabels (Sec.~\ref{subsubsec:modelannots}) from the best model from the previous stage. We train on these new annotations and report our results in Tab~\ref{tab:exp_pseudolabels}. When we use the refined detections but still use the mouthing cues for assigning word labels, we obtain a \namecerfspell{} of 57.5, a similar result to the corresponding model (57.6) from Stage 1. 

\noindent\textbf{\modelname{} letter labels.}
The error rates drop further when we also improve the recognition annotations. We do so by using a variant of the edit distance to match the predictions of the Stage~1 Transpeller to a word in a neighboring subtitle. These results can be seen in row 2 of Tab.~\ref{tab:exp_pseudolabels}. 

\noindent\textbf{Finetuning with MH-CTC.}
Finally, when finetuning further with our MH-CTC loss, our final Stage 2 model gives a \namecerfspell{} of 53.3 which is 4.3 points better than the best model (57.6) of Stage 1. It is evident from both these tables that both MH-CTC and our multi-stage training with pseudolabelling improve our recognition performance.

\noindent\textbf{Fingerspelling recognition vs. spotting mouthings.} An interesting line of thought is: if we have access to the subtitles at test time, can we use a mouthing model to accurately predict words instead of doing fingerspelling recognition? To judge this, we restrict the test set to instances where there is a noun or a proper noun mouthing annotation and compute the CER between the mouthing annotation and the ground truth annotation, and we get 55.6. On this same subset, our best model obtains 53.3 CER, demonstrating that it in fact performs better than a mouthing method with access to the subtitle text.

\begin{figure}
\begin{floatrow}
\ffigbox{%
  \includegraphics[width=0.43\textwidth]{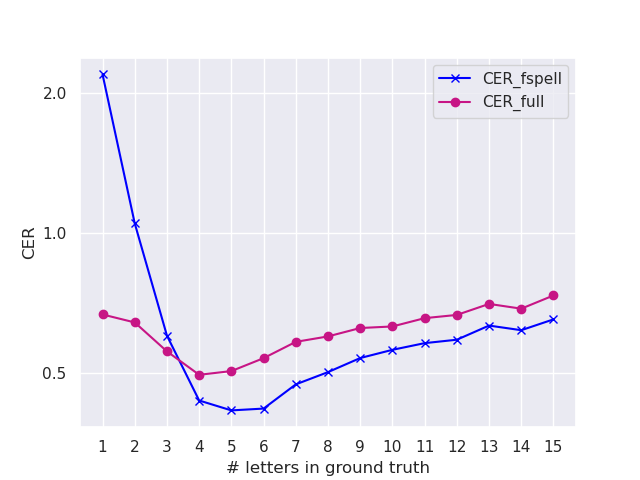}
  
}{%
  \caption{\textbf{Variation of CER vs the number of letters in the ground-truth.} Transpeller struggles to correctly predict very short fingerspelling segments, which are usually partial fingerspellings (e.g. MJ for Mary Jane). This happens because the model is only supervised with full-length words during training.}%
      \vspace{-0.3cm}
  \label{fig:cer_with_length}
}

\capbtabbox{%
  \begin{tabular}{lll}
        \toprule
        %{\footnotesize
        \textbf{Ground Truth} & \textbf{Prediction} & \textbf{CER} \\\midrule
        chloride & churide & 25 \\
        nurembrg  & turmug & 50  \\
        ivory & ener & 80  \\
        elind & elinc &  20 \\
        bnmm & ben & 75 \\
        clove & clune & 40 \\
        semnoa & samol & 50 \\
        % virgosimonkk & vrgesamon & 42  \\ 
        %nut  & lat & 67 \\
       % }
        \bottomrule
    \end{tabular}
}{%
  \caption{\textbf{Qualitative examples.} Model predictions on our manually verified test set. The CERs are shown for reference. We see that one of the most common error sources is confusion between letters that are visually similar: (a, e, i), (d, c), (l, n, m, v, t), (o, u).}%
      \vspace{-0.3cm}
  \label{tab:qualitative_testset}
}
\end{floatrow}
\end{figure}

\noindent\textbf{Error analysis of the best model on the test set.} 
% Restricting the test set to instances where there is a noun or a proper noun mouthing annotation, we can evaluate the CER between the mouthing annotation and the ground truth annotation, which is 48.3. On this subset of the test set, our best model obtains a CER of 51.3, demonstrating that it is performing almost as well as an `oracle' method with access to the subtitle text. 
In Tab.~\ref{tab:qualitative_testset}, we show a few examples of our predictions and the corresponding ground-truth sequence. We can see our model makes reasonable errors for most examples, where it confuses between letters that are visually quite similar. In Fig.~\ref{fig:cer_with_length}, we show how the CER varies based on the length of the ground-truth character sequence. It is evident that the model struggles the most with very short sequences (one or two letters). This is expected because it has only been supervised with full word annotations during training and has never been trained to predict one or two letters in isolation.

\noindent\textbf{Lookup-based correction at inference-time.} We explore the possibility of correcting the errors in the model's outputs at test-time with the help of a pre-determined list of words, e.g. ``atlanic" to ``atlan\textbf{t}ic". 
We first curate a list of nouns present in the subtitles of the BOBSL train set and use edit distance to match the model's predictions to the closest noun in our list. If the edit distance is below a set threshold, i.e., a very close match, we replace the predicted character sequence with the matched word from our list. 
However, we found that this increases the \namecerfspell{} to 57.0 and \namecerfull{} to 61.1. Such a lookup-based correction method leads to several false matches because (i) it is done with no context of the surrounding words, (ii) not all the letters of the word are fingerspelled, (iii) fingerspelled words in the test set can also be novel and unseen.

\vspace{-0.3cm}
\subsection{Architecture ablations}
\vspace{-0.2cm}

\noindent\textbf{Importance of joint recognition and detection.} As described in Sec~\ref{subsec:model}, our model contains three prediction heads for classification, localization, and recognition. 
%(i) a classification head that predicts if a given video clip contains fingerspelling, (ii) a localization head to detect the temporal location of the fingerspelling, and (iii) the recognition head to determine the letters that are fingerspelled. 
All three heads are essential and are used to obtain the Stage 2 annotations as described in~\ref{subsubsec:modelannots}. We conduct an experiment to also demonstrate that it is beneficial to train these heads jointly. We find that joint training leads to a better recognition performance (55.4 \namecerfspell{}) than training without the localization head (56.3 \namecerfspell{}) or without the localization and classification heads (56.2 \namecerfspell{}). 

\noindent\textbf{Sequence-to-Sequence vs. CTC-based models.}
We also compare our CTC-based recognition head with a sequence-to-sequence (seq2seq) encoder-decoder architecture supervised with a cross-entropy loss. We use the standard Transformer-Base~\cite{vaswani2017attention} model, which contains a Transformer encoder similar to Transpeller and a 6-layer auto-regressive Transformer decoder. We compare with the CTC model trained on the Stage 2 annotations. 
%Both models decode the character sequence with a beam size of 30 during inference. 
The seq2seq network performs worse (57.4 \namecerfspell{}) than the CTC model (55.4 \namecerfspell{}). This is expected because only 22\% of the test samples contain no missing letters, i.e. full words, so conditioning on past letters can lead to errors in future letter predictions. However, conditioning on past letters can be very helpful if we want to actually estimate the full word and not determine the exact letters that are fingerspelled. This is indeed validated by the fact that the seq2seq model performs much better on \namecerfull{}, achieving 55.2   compared to the CTC model's 63.0. We also note that due to the sequential decoding in the seq2seq model, it is $\approx20\times$ slower in terms of run-time speed than the CTC model, which decodes all the characters in parallel.

\begin{figure}[t]
    \centering
    \includegraphics[width=\textwidth]{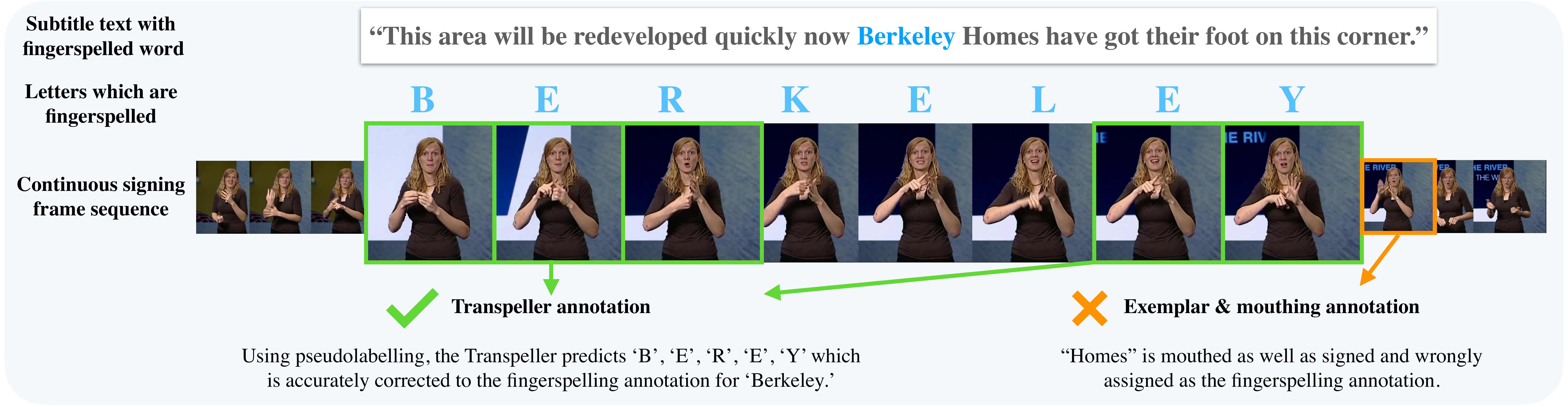}
    \caption{\textbf{Benefit of Transpeller annotations:} Here, we are given a fingerspelling clip of ``BERKELEY''. The initial annotation with exemplars + mouthing cues assigns an incorrect word label because the spotted mouthing of the word ``Homes'' is temporally close to the fingerspelling location, resulting in a false word assignment. In the second annotation stage, we correct this using the Transpeller's predicted characters ``BEREY'' that are matched to the subtitle word ``BERKELEY''.
    }
        \vspace{-0.3cm}
    \label{fig:error}
\end{figure}

% \begin{figure}[t]
%     \centering
%     \includegraphics[width=0.5\textwidth]{cer_by_number_gt_chars.png}
%     \caption{\textbf{Variation of CER vs the number of letters in the ground-truth.} Transpeller struggles to correctly predict single letter or double letter fingerspelling segments, which are usually partial fingerspellings (e.g. MJ for Mary Jane). This happens because the model is only supervised with full-length words during training.
%     }
%     \label{fig:cer_with_length}
% \end{figure}

%\vspace{-10pt}

\vspace{-0.4cm}
\section{Conclusion}
\label{sec:conclusion}
\vspace{-0.2cm}

We presented a new BSL fingerspelling recognition benchmark and our Transpeller model
designed to jointly detect and recognize fingerspelled letters in continuous sign language video.
Our training data is largely constructed automatically, exploiting English subtitles and mouthing cues. The evaluation data is manually curated, and we achieve promising recognition results (Tab.~\ref{tab:qualitative_testset}).
% However, there is room for improvement, especially to incorporate higher resolution inputs around hand regions (as in~\cite{shi2021fingerspelling}).
% While Transpeller obtains promising results,
However, we note some limitations. 
First, there remains room for improvement in the accuracy of the model to further reduce the character error rate.
Second, our training and evaluation of the Transpeller is limited to the use of interpreted data and therefore is not necessarily representative of more natural, conversational signing.
Addressing these limitations would be a valuable future research direction. 
% Future work can further explore end-to-end training to obtain suitable video representations
% specific to fingerspelling.

\subsubsection*{Acknowledgements} 
This work was supported by the Oxford-Google DeepMind Graduate Scholarship, EPSRC grant ExTol, a Royal Society Research Professorship and the ANR project CorVis ANR-21-CE23-0003-01. HB would like to thank Annelies Braffort and Michèle Gouiffès for the support.

\bibliography{references}

% \newpage
\bigskip
{\noindent \large \bf {APPENDIX}}\\
\renewcommand{\thefigure}{A.\arabic{figure}} % \thesection instead of A would make it A.1, B.1...
\setcounter{figure}{0} 
\renewcommand{\thetable}{A.\arabic{table}}
\setcounter{table}{0} 

\appendix

%%%% Example usage of referencing to the main paper

% As demonstrated in
% \if\sepappendix1{Tab.~1}
% \else{Tab.~\ref{tab:cosinesimilaritystats}}
% \fi
% ...

%%%%

We provide further details on the computation of feature similarity to obtain fingerspelling localisations in Sec.~\ref{sec:features_similarity}. Details on our modified text edit distance are in Sec.~\ref{sec:edit_dist}. Implementation details on the Transpotter model are provided in Sec.~\ref{sec:implementation_details} and an experiment demonstrating our choice of features is recorded in Sec.~\ref{sec:choice_feats}. Finally, we show qualitative examples of our results on the test set as well as our pseudolabels on the training set in Sec.~\ref{sec:qualitative_test}. 

\vspace{-0.5cm}
\section{Video results} We highly recommend checking out the video results at:\\ \url{https://www.robots.ox.ac.uk/~vgg/research/transpeller/}. 

\vspace{-0.5cm}
\section{Feature similarity}
\label{sec:features_similarity}

As mentioned in
\if\sepappendix1{Sec.~4.1,}
\else{Sec.~\ref{subsubsec:exemplarannots},}
\fi
we use feature similarity to search for fingerspelling instances, given a small number of manual annotations ($115$ examples of frames containing fingerspelling). We compute the cosine similarity between these exemplars $E$ and features of the videos in our corpus. 

Formally, given a video $V$ with $N$ frames and $d$ dimensional features corresponding to each frame, we can express $V$ as a $N\times d$ matrix, which we call $\mathcal{V}$. Similarly, we can express the $E$ exemplar frames as a $E\times d$ matrix $\mathcal{W}$. The cosine similarity between rows of $\mathcal{V}$ and rows of $\mathcal{W}$ is a measure of similarity between frames of $V$ and fingerspelling. We can thus normalise $\mathcal{V}$ and $\mathcal{W}$ along the feature dimension, denoted $\mathcal{V}_*$ and  $\mathcal{W}_*$, and compute a similarity matrix $\mathcal{S} = \mathcal{V}_*\times \mathcal{W}_*^T$ of dimension $N \times E$. We consider the percentage of columns $c$ of $\mathcal{S}$ with a similarity value about a threshold $S$. If $c$ is greater than a threshold $C$, then we consider the frame to be fingerspelling. In our case, we let $S=0.4$ and $C=0.2$. We apply some smoothing to the edges of the fingerspelling detections and remove very short segments of either fingerspelling or gaps between fingerspelling localisations. 
%Statistics on the number and duration of these fingerspelling detections can be found in Tab.~\ref{tab:cosinesimilaritystats}.

\vspace{-0.5cm}
\section{Text proximity score}
\label{sec:edit_dist}

In order to pseudo-label the training dataset using the outputs of the Transpeller model from Stage~1, we find potential fingerspelled words in the subtitles with a high proximity score to the decoded outputs, as described in
\if\sepappendix1{Sec.~4.2.}
\else{Sec.~\ref{subsubsec:modelannots}.}
\fi
This proximity score is a modified version of the Levenshtein edit distance, as the classic Levenshtein edit distance between two strings is not necessarily well adapted to our task of fingerspelling recognition. Although the first letter of a word is almost always fingerspelled, other letters may be omitted, e.g.\ `Scapa Flow' may be fingerspelled `SCAPAFW' and `HARRY' may be fingerspelled `HRY'. Letting $w_1$ be the subtitle word and $w_2$ be the Transpeller outputs, we compute our adapted edit distance dist($w_1$, $w_2$) as follows: 

\begin{enumerate}
    \item All repeated letters in $w_1$, $w_2$ are removed before computation, e.g. (HARRY becomes HARY). 
    \item Letters in $w_2$ that are not in $w_1$ are penalised and removed, e.g. (HARY, HERY) becomes (HARY, HRY) with a malus of +1. 
    \item Correct prediction of the first letter reduces the edit distance by 1, e.g. (HARY, HRY) has a bonus of -1, but (HARY, ARY) does not.
    \item Insertions and subtitions each carry a malus of +1, e.g. (HARY, HYR) has a malus of +1.
    \item Deletions are not penalised.
    \item The proportion of letters of $w_1$ not in $w_2$ is added as a fractional malus, e.g. (HARY, HRY) has a malus of +1/4. This is to ensure that dist(HARY, HRY)$<$dist(HUMPHREY, HRY), given that deletions are not penalised. 
\end{enumerate}

% We compare our modified edit distance to the standard Levenshtein edit distance. 

% \begin{table}
% \centering
% \begin{tabular}{llll}
% \toprule
% \textbf{Annotations} & \textbf{\# Recognition ex.} & \textbf{\namecerfspell} & \textbf{\namecerfull} \\
% \midrule
% \modelname{} detect. + Char. labels: Lev. dist.    & 111k  &      &       \\
% \modelname{} detect. + Char. labels: our dist. & 111k & 52.4 & 55.4 \\\bottomrule
% \end{tabular}
% \caption{\textbf{Levenshtein edit distance vs. our modified edit distance.} Due to the characteristics of fingerspelling, particularly the frequent dropping of letters, our modified edit distance is a better measure of proximity between words from the subtitle and Transpeller outputs. }
% \label{tab:edit_distance_exp}
% \end{table}

\section{Implementation details}
\label{sec:implementation_details}

\noindent\textbf{Data sampling.} We form the training samples $\mathcal{D}$ by randomly (50\% chance) sampling positive (with fingerspelling) or negative (without fingerspelling) clips. Note that we have a lot more detections of fingerspelling (see
\if\sepappendix1{Tab.~1),}
\else{Tab.~\ref{tab:cosinesimilaritystats}),}
\fi
but only a fraction of them are associated with a word label. The duration of the sampled clips ranges from 1.2s to 4.8s. For training the Stage 1 model, we found that a simple augmentation strategy such as randomly dropping characters (except the first character) helps combat overfitting in the low data regime. This was, however, not helpful for our Stage 2 model training, when we obtain a larger training set.

\noindent\textbf{Data splits.} All our models are trained on the automatic annotations for the videos in the \textit{train} split of the BOBSL dataset. Similarly, for validation, we use the automatic annotations of the videos in the \textit{val} split. We choose the checkpoint with the lowest validation loss and evaluate it on the manually verified test benchmark

\noindent\textbf{Automatic annotations.} We use Flair \cite{akbik2018coling} parts of speech tagging to identify nouns and proper nouns in the subtitle text. We take mouthing annotations with confidence scores of above 0.1, using the model in \cite{Momeni22}. 

\noindent\textbf{Evaluation.} At evaluation, we give the model as input the context window seen by the annotators, i.e. 2.1s before and 4s after the midpoint of the automatically detected fingerspelling instance. 

\noindent\textbf{Lookup-based correction at inference-time.} We first curate a list of nouns present in the subtitles of the BOBSL train set and use edit distance to match the model's predictions to the closest noun in our list. If the edit distance is below a set threshold, i.e., a very close match, we replace the predicted character sequence with the matched word from our list.

\section{Choice of features}
\label{sec:choice_feats}

To extract the visual features, we use a Video-Swin-S model pretrained for the sign classification task. Previous works~\cite{bull2021aligning,Momeni22,varol2021read} used an I3D model in a similar fashion.  In Tab.~\ref{tab:i3d_swin}, we compare these two feature extractors and show that our Video-Swin-S features are far superior to the I3D features.

\begin{table}[h]
\centering
\begin{tabular}{llll}
\toprule
\textbf{Annotations} & \textbf{\# Recognition ex.} & \textbf{\namecerfspell} & \textbf{\namecerfull} \\
\midrule
I3D (Exemplars + Mouthings: proper nouns)    & 39K  & 64.6     &   66.5    \\
Swin (Exemplars + Mouthings: proper nouns) & 39K & \textbf{58.5} & \textbf{62.1} \\\bottomrule
\end{tabular}
\caption{\textbf{I3D vs.~Swin features.} Using the Video-Swin-S model rather than an I3D model trained for sign classification provides markedly better features. We thus train the Transpotter model on Swin features in all of our experiments. }
\label{tab:i3d_swin}
\end{table}

\section{Qualitative examples}
\label{sec:qualitative_test}

\begin{figure}[t]
    \centering
    \includegraphics[width=\textwidth]{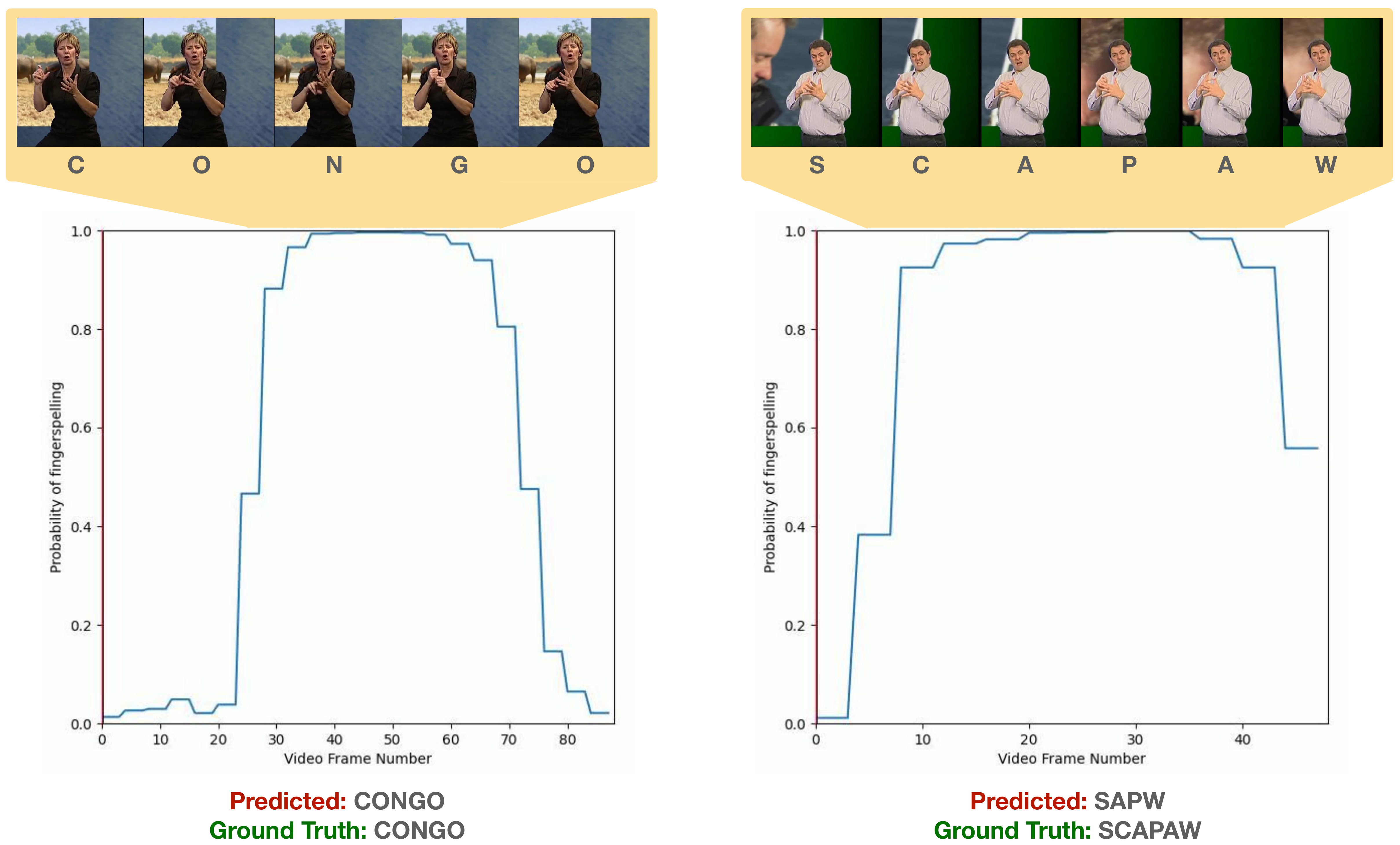}\vspace{5pt}
    \includegraphics[width=\textwidth]{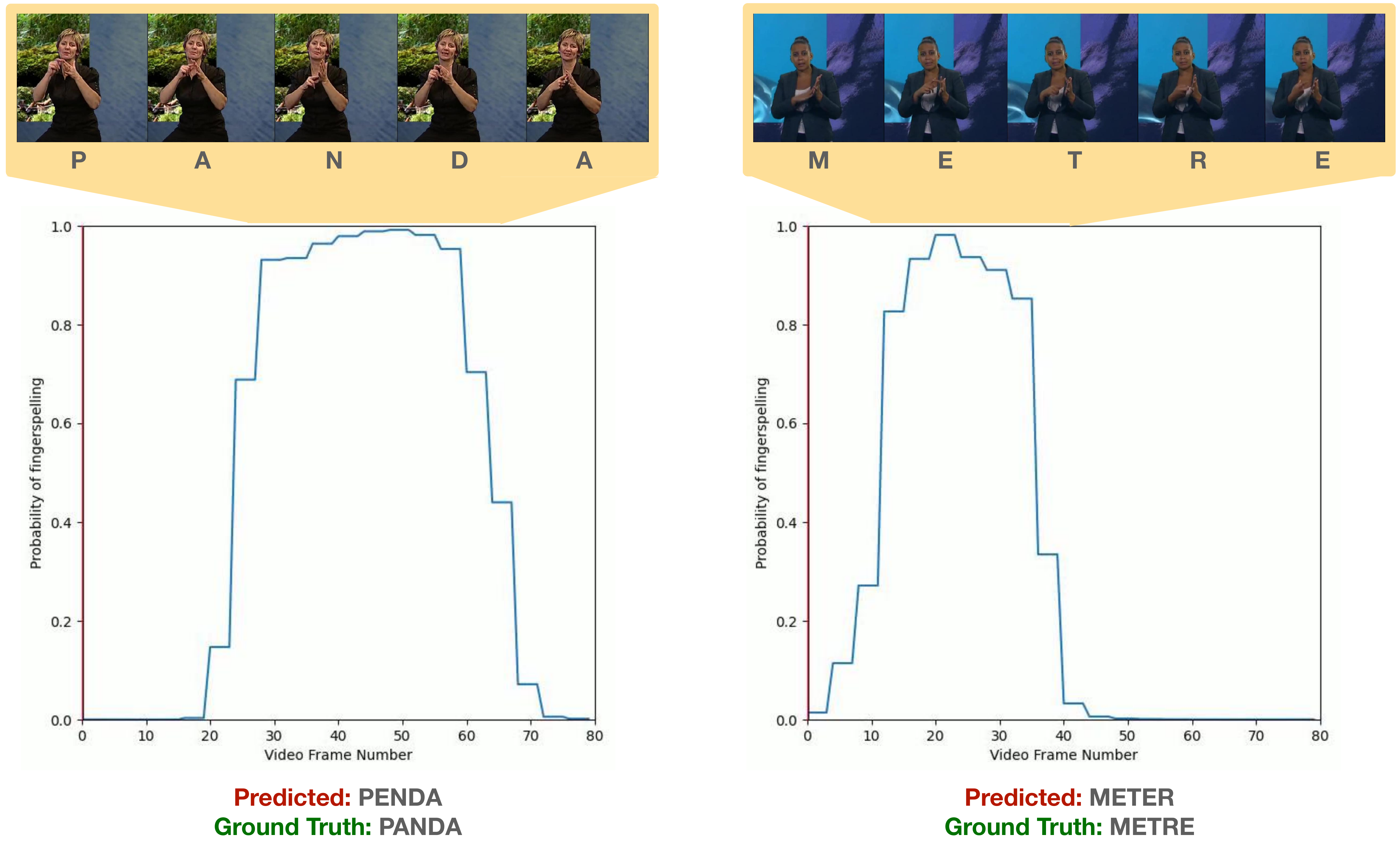}
    \caption{\textbf{Success and failure cases on the test set.} In the top left example, our model correctly predicts the fingerspelling letter sequence. The top right example shows how our model fails to recognise some letters when the fingerspelling is fast or when the hand movements are blurry. In the bottom left example, our model confuses the letter `E' and `A', which are very similar in BSL. Finally, the bottom right example shows a case where the model switches the order of two letters. Alternatively, perhaps our model is too reliant on a language model it has learned from the training examples, and predicts an alternative spelling of `METRE'. 
    }
    \label{fig:app:testset_ex}
\end{figure}

\begin{figure}[t]
    \centering
    \includegraphics[width=\textwidth]{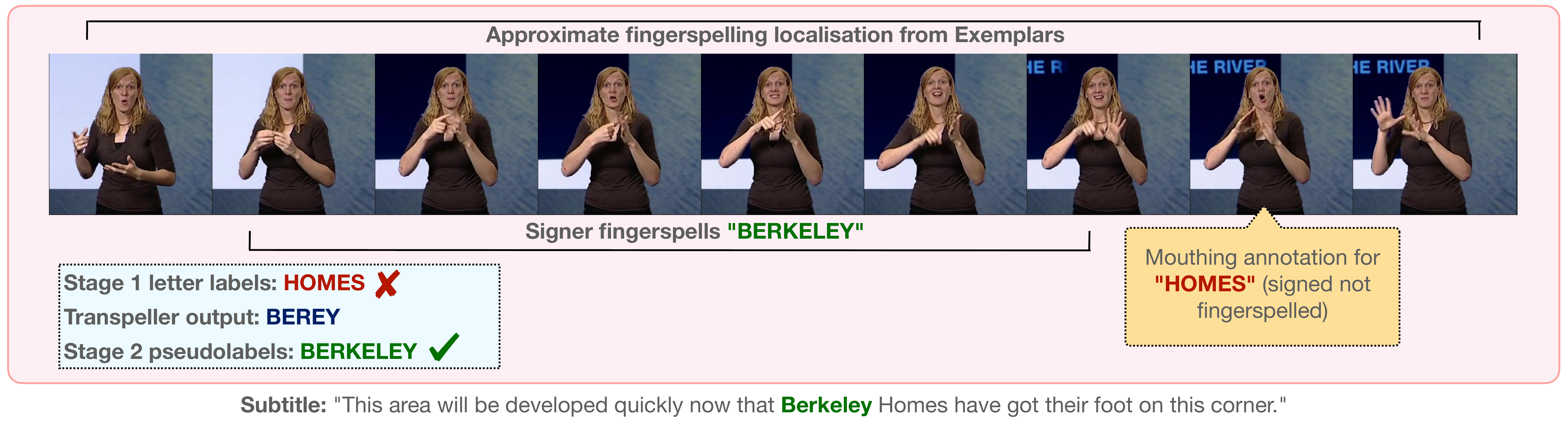}\vspace{5pt}
    \includegraphics[width=\textwidth]{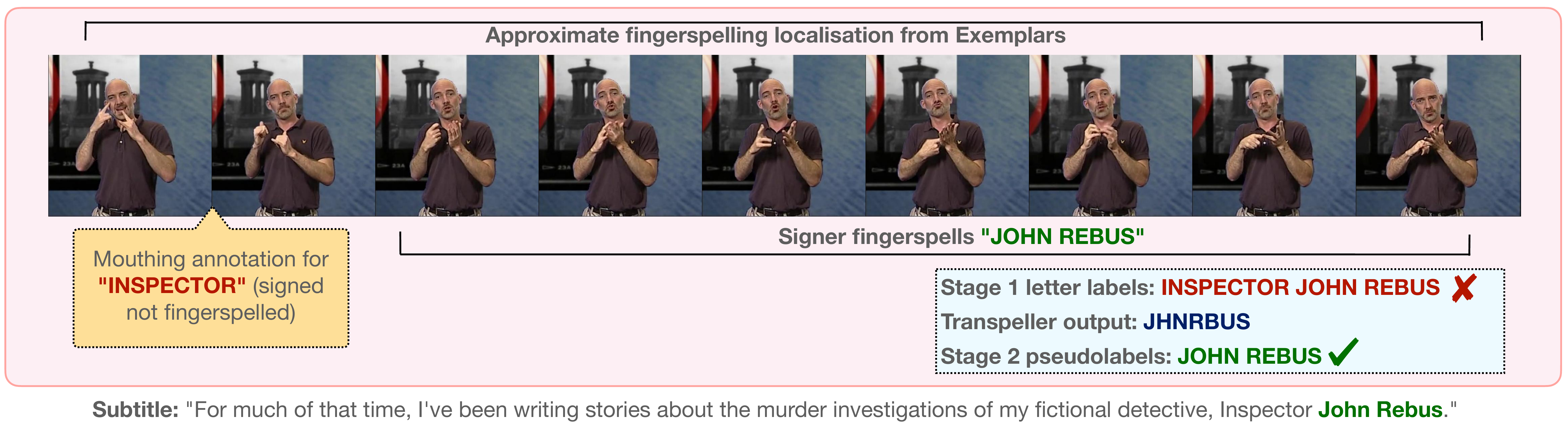}\vspace{5pt}
    \includegraphics[width=\textwidth]{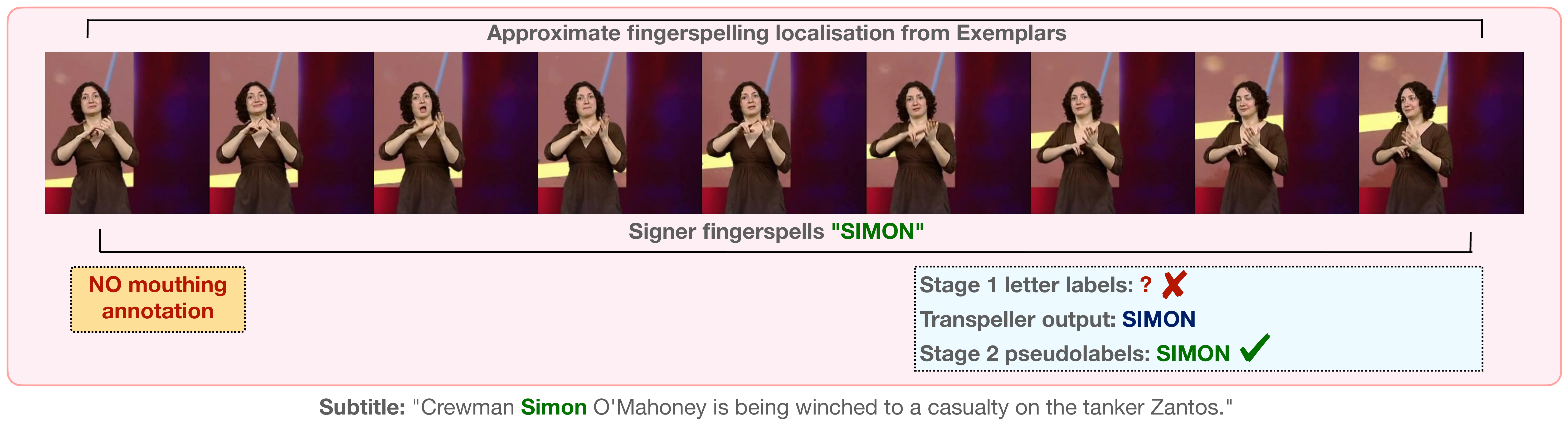}
    \caption{\textbf{Improvements to our pseudolabels in Stage 2.} During Stage 1, we use mouthing cues to annotate fingerspelling. These mouthing cues may not correspond to the word which is being fingerspelled, but rather a sign either directly before or directly after the fingerspelling. The boundaries of our fingerspelling localisations are not necessarily accurate enough to exclude these mouthings. In other cases, the mouthing model fails to accurately spot a mouthing cue in a fingerspelling segment, and so we do not have an annotation. Using the Transpeller outputs in combination with the subtitle text, we are able to improve our training annotations for Stage 2.  
    }
    \label{fig:app:improving_pslabs}
\end{figure}

Fig.~\ref{fig:app:testset_ex} shows examples of success and failure cases of the Transpotter model on our test set. Fig.~\ref{fig:app:improving_pslabs} demonstrates how letter labels from mouthings in Stage 1 are improved by Transpeller pseudolabels in Stage 2 for better supervision. 

\end{document}